%% file: acl2021.tex
\documentclass[11pt,a4paper]{article}
\usepackage[hyperref]{acl2021}
\usepackage{times}
\usepackage{latexsym}
\usepackage{tabularx}
\usepackage{color,soul}
\usepackage{multirow}

\usepackage{microtype}
\usepackage{graphicx}
\usepackage{amsmath}

\aclfinalcopy 
\def\aclpaperid{4023} 

\title{Towards a Universal NLG for Dialogue Systems and Simulators with Future Bridging}

\author{Philipp Ennen$^1$ \\ MediaTek Research \And Yen-Ting Lin \\ MediaTek Research
  \And Ali Girayhan Özbay \\ Imperial College London \And Ferdinando Insalata \\ MediaTek Research \AND Maolin Li \\ MediaTek Research \And Ye Tian \\ MediaTek Research \And Sepehr Jalali \\ MediaTek Research \And Da-shan Shiu$^1$ \\ MediaTek Research \\ \AND \\
  $^1$\texttt{\{philipp.ennen, ds.shiu\}@mtkresearch.com } }

\date{}

\begin{document}
\maketitle

\input{paper/abstract.tex}
\input{paper/introduction}
\input{paper/related_work}
\input{paper/ductiletransformer}
\input{paper/evaluation}
\section{Discussions and Future Work}

\textbf{Accepting a further-away future}. In this work, the evaluated FBNLG prototype takes into account a future that is the desired user utterance in the next turn. We regard the most critical immediate extension of this work to have an FBNLG take a target that is a mixture of utterances that are further into the future. \\

\noindent\textbf{Fully scaling up prototype}. In this work, our prototype was built by fine-tuning an existing language model with very modest data. We are motivated to see the performance of an FBNLG pretrained using e.g. the 341 GB of text used to train Meena \cite{adiwardana2020towards}. With a properly built FBNLG, one then can properly examine the zero-shot and few-shot properties. \\

\noindent\textbf{Interworking with compatible components}.  
An obvious future work is to demonstrate the full end-to-end behaviour of a compatible stack on top of an FBNLG. Also, to be useful in a real-world scenario, a dialogue system cannot return false information to a user. Thus a future extension of this work is to apply knowledge and fact grounding techniques to augment an FBNLG.

\section{Conclusion}
In this paper, we proposed a future bridging NLG concept. The concept enables a universal NLG for dialogue systems and simulators using self-supervised training from massive datasets. An FBNLG is expected to support a wide variety of domains and dialogue behaviours with few-shot or even zero-shot adaptation. To demonstrate the feasibility of the FBNLG approach, we constructed a prototype by fine-tuning a large language model. The results indicated that FBNLG can generate system responses that are reasonably fluent, specific, sensible, even for never-planned dialogue behaviours such as dialogue type switching without any learning.



\bibliographystyle{acl_natbib}
\bibliography{acl2021}
\end{document}


\maketitle

\appendix




\section{Appendices}
\label{sec:appendix}
\begin{table*}[t!]
\centering
\begin{tabularx}{1.0\textwidth}{X|l|lllllll}

\hline \textbf{Dialogue type transition} & \textbf{Metric type} & \textbf{1} & \textbf{2} & \textbf{3}  & \textbf{4} & \textbf{5} & \textbf{6}  & \textbf{7}\\ \hline
chit-chat to task-oriented for future-bridging & Specificity & 0.14 & 0.10 & 0.14 & 0.16 & 0.16 & 0.18 & 0.12 \\
  & Sensibleness & 0.07 & 0.06 & 0.08 & 0.12 & 0.19 & 0.21 & 0.27 \\ \hline
chit-chat to task-oriented for dialogue-type transition & Specificity & 0.15 & 0.11 & 0.14 & 0.12 & 0.20 & 0.17 & 0.11 \\
  & Sensibleness & 0.03 & 0.04 & 0.07 & 0.10 & 0.22 & 0.24 & 0.30 \\ \hline
task-oriented to chit-chat for dialogue-type transition & Specificity & 0.02 & 0.04 & 0.10 & 0.15 & 0.13 & 0.19 & 0.37 \\
  & Sensibleness & 0.02 & 0.05 & 0.15 & 0.16 & 0.17 & 0.18 & 0.27 \\ \hline
task-oriented to chit-chat for future-bridging & Specificity & 0.02 & 0.04 & 0.10 & 0.15 & 0.13 & 0.19 & 0.37 \\
  & Sensibleness & 0.02 & 0.05 & 0.15 & 0.16 & 0.17 & 0.18 & 0.27 \\ \hline
\end{tabularx}
\caption{The distribution of ratings for each experiments. The rating is on the Likert-scale between 1 (very bad) and 7 (very good).}
\label{tab:human_eval_inter}
\end{table*}
\subsection{Human evaluation setup}
For each experiment, we asked 20 workers to rate 20 dialogues, respectively. For each dialogue, we asked the workers to evaluate sensibleness and specificity of the generated system response on a Likert scale between 1 (very bad) and 7 (very good). For this, we have presented dialogues that are created from a fused test dataset of SGD and DailyDialog to the workers. A dialogue in this fused dataset is a combination of a chit-chat conversation coming from DailyDialog followed by a task-oriented one drawn from SGD or vice versa. \\

\noindent Table \ref{tab:human_eval_inter} gives the distribution of ratings for each experiment as presented in Table 4 and 5 of the main paper. 

\subsection{Special Tokens}\label{sc:special_tokens}
\noindent Prior work have shown that language models benefit from special tokens \cite{peng2020soloist,hosseini2020simple}. Special tokens provide structure for the model input by adding hints of what segments within the input are about. 
Accordingly, we have added special tokens for user turns (``USER"), system turns (``SYSTEM") as well as beginning of sentence (``BOS") and end of sentence (``EOS") for the beginning and the end of the entire input sequence respectively. In order to make our model robust against the absence and presence of future, we do not introduce a dedicated token to represent the future. Instead, in our experiments, given that we use the desired user utterance at one turn away as the future, we simply used the user turn token ``USER" to indicate the start of the future. 

\bibliographystyle{acl_natbib}
\bibliography{acl2021}

%% file: paper/abstract.tex
\begin{abstract}
In a dialogue system pipeline, a natural language generation (NLG) unit converts the dialogue direction and content to a corresponding natural language realization. A recent trend for dialogue systems is to first pre-train on large datasets and then fine-tune in a supervised manner using datasets annotated with application-specific features. Though novel behaviours can be learned from custom annotation, the required effort severely bounds the quantity of the training set, and the application-specific nature limits the reuse. In light of the recent success of \textit{data-driven} approaches, we propose the novel \textit{future bridging} NLG (FBNLG) concept for dialogue systems and simulators. The critical step is for an FBNLG to accept a \textit{future} user or system utterance to bridge the present context towards. Future bridging enables self supervised training over annotation-free datasets, decoupled the training of NLG from the rest of the system. An FBNLG, pre-trained with massive datasets, is expected to apply in classical or new dialogue scenarios with minimal adaptation effort. We evaluate a prototype FBNLG to show that future bridging can be a viable approach to a universal few-shot NLG for task-oriented and chit-chat dialogues.
\end{abstract}

%% file: paper/introduction.tex
\section{Introduction}

In a dialogue system pipeline, after the information from a user is extracted, the dialogue direction and the information content to return to the user for the immediate turn is decided. This is followed by a unit referred to as natural language generation (NLG) converting the information content to a corresponding natural language realization. A recent trend is toward building pipelines with largely machine learning methods.  \\

\noindent Learning-based components in a dialogue system are typically trained in a supervised manner using sample dialogues annotated with application-specific features. In some systems, certain components are pre-trained first by massive datasets \cite{conversational_gpt_huggingface,henderson-etal-2019-training,golovanov-etal-2019-large}. The application-specific annotation is critical for a dialogue system to learn its functionality. For instance, annotations are necessary for task-oriented dialogues to learn to extract dialogue states \cite{10.1145/357417.357420,budzianowski2018multiwoz,shah2018}. An open-domain dialogue system often learns to compute similarity scores for response retrieval from carefully curated datasets \cite{jafarpour2010filter,Leuski_Traum_2011,ji2014information,boussaha2019deep}. \\

\noindent Unfortunately, the required annotation effort severely limits the quantity of the training set, which in turn limits the system performance. Furthermore, because a training set is annotated with domain-specific features, a trained pipeline can only function well in the corresponding narrow domain. This makes it difficult to directly deploy or at least transfer a trained pipeline to new domains \cite{dstc8} or to new dialogue behaviours, such as goal-guided topic traversal \cite{tang-etal-2019-target}. \\

\noindent By not requiring any additional annotation, self-supervised training allows several orders of magnitude increase of the training set size compared to supervised training. This has been shown to lead to not only higher performance but versatility in the application domain, and also ease of creating never-planned dialogue behaviours. For instance, GPT-2 \cite{gpt2} can generate coherent text for various topics and styles simply from priming. Gshard \cite{lepikhin2020gshard}, trained using 25 billion training examples, obtained far superior quality for translation from 100 languages to English compared to the prior art. \\

\noindent In this paper, we aim to build a \textit{universal} NLG. This NLG can be used in various dialogue types, for intended and never-planned behaviours, with zero-shot or few-shot adaptation. In light of the success of these \textit{self-supervised} approaches, we surmise that to achieve this, one shall decouple the training of this NLG from the rest of a dialogue system, and pre-train the NLG over a large and general dataset in a self-supervised manner. \\

\noindent To that end, we propose the critical concept, \textit{future bridging} (FB). Future bridging is key to enable decoupled, self-supervised training of an NLG. The text-infilling formulation of future bridging enables the easy collection of a large-sized, annotation-free dataset. An FBNLG takes not only the present dialogue context but also a \textit{future} user or system utterance as input. It shall then predict the text that has a high probability of bridging the context to that desired future utterance. In this set up, the top stack of a dialogue system can select a future from one corpus, while the NLG can learn the act of bridging from a different corpus. Thus the training of the NLG is decoupled. \\

\noindent To better explain the future bridging concept, two hypothetical dialogues are given in Table 1. Here, an FBNLG is given the desired next-turn user utterance as the future to bridge forward to. We note that there is a very intuitive and satisfying interpretation for future bridging. If a user utterance is selected as the target future, such utterance represents what the system wishes to induce the user to ultimately say. On the other hand, if a system utterance is selected as the target future, that utterance represents what the system wishes the dialogue to develop towards so that the system is in position to say that. \\

\begin{table}[t!]
\small
\textbf{Dialogue 1:} Example dialogue of task-oriented type \\
\begin{tabularx}{0.48\textwidth}{|lX|}
\hline
\textbf{User} & Hi Jarvice. I want to buy a movie ticket. \\
\textcolor{gray}{\textbf{\textit{Projected future}}} & \textcolor{gray}{\textit{I want to see Toy Stories Eight.}} \\
\textbf{FBNLG} & Do you want to see action, or comedy, or some animation? \\
\textbf{User} & I'd like to watch some animation. \\
\hline
\end{tabularx}

\vspace{0.4em}

\textbf{Dialogue 2:} Example dialogue from chit-chat to persuasion \\
\begin{tabularx}{0.48\textwidth}{|lX|}
\hline
\textcolor{gray}{\textbf{\textit{Projected future}}} & \textcolor{gray}{\textit{NULL}} \\
\textbf{FBNLG} & Howdy! I heard the weather's gonna be good this weekend. \\
\textbf{User} & Yup. Uh, I want to buy a movie ticket. \\
\textcolor{gray}{\textbf{\textit{Projected future}}} & \textcolor{gray}{\textit{I'd like that Thai restaurant}.} \\
\textbf{FBNLG} & The traffic is kind of bad out there to go to any movie theater. Can I interest you with a few nearby restaurants, like King of Burgers or Best of Thai, instead? \\
\textbf{User} & OK, show me Burgers. \\
\hline
\end{tabularx}
\caption{Illustrating examples of the \textit{future projection} and the \textit{future bridging} concept. (a) A typical task-oriented dialogue. (b) Start with greeting then move to a persuasive dialogue behaviour to fulfill a hidden system goal. Persuasion is not a typical scenario in a task-oriented dataset.}
\label{tab:example_dialogues}
\end{table}
\noindent We remark that in both example dialogues, the user never actually produce the supplied future utterances. This highlights the very nature of the \textit{future}; it serves to indicate the direction, instead of the landing spot, for the NLG to shoot for. \\

\noindent An important benefit from adopting a \textit{future bridging} interface is that it can greatly simplify the training of a dialogue policy via reinforcement learning (RL). Many studies \cite{Asri+2016,Crook2017,li2017investigation,li2017user} employ RL to learn a policy from experiments. Both a user simulator and a system simulator are required. However, building a human-like, high diversity natural language generator component for either simulator is very difficult. As a result, studies may resort to using rule-based response templates, or even just skip the natural language realm entirely \cite{shah2018}. In our opinion, a universal NLG such as a FBNLG can vastly simplify the current challenge in dialogue policy learning.  \\

\noindent We claim our contributions as follows. 
\begin{itemize}
    \item We propose the future bridging concept. The concept allows a FBNLG to be self-supervised trained over a massive dataset, decoupled from the rest of system. 
    \item An FBNLG is expected to support a wide variety of domains and dialogue behaviours with zero-shot or few-shot adaptation.
    \item A FBNLG can be used in a simulator to greatly reduce the difficulties of learning a dialogue policy via RL.
    \item We evaluated a prototype FBNLG and showed that it could indeed support a novel use of case supporting both classical and a novel use case of seamlessly transitioning between chit-chat and task-oriented types. 
\end{itemize}

\if false

\section{Introduction}

In a dialogue system pipeline, after relevant information from a user is extracted, a dialogue management (DM) unit decides on the direction and the information content to return to the user. This is followed by another component natural language generation (NLG) unit converting the information content to a corresponding natural language realization. A recent trend is towards building these two components jointly with largely machine learning methods. Such a system is trained in a supervised manner from dialogues dedicated to a well-defined domain. By incorporating specialized annotations, some systems learn to vary the topics of ongoing dialogues. In this paper, we present DuctileGPT, a novel transformer-based NLG method which aims to decouple the dialogue direction/fact and knowledge determination step and the surface realization step for dialogue systems.  

\noindent Research in automated dialogue systems is often demarcated between task-oriented dialogues and chit-chat. Task-oriented dialogue systems tend to involve heavy engineering of the dialogue policy, `intents and slots' based approach for Natural Language Understanding (NLU), and template based approaches for Natural Language Generation (NLG) \cite{gao2018neural}. In this approach, the dialogue system first tries to identify the task requested by the user and then tries to establish pieces of key information necessary to accomplish that task (i.e. filling in the `slots'), either by analyzing the initial prompt by the user or by asking further questions. As an example, a task-oriented dialogue system may be prompted by the user to order takeaway delivery from a particular restaurant, and as a response, the dialogue system may ask which items from the menu the user would like and where the food should be delivered. 

\noindent Despite their success, most such systems extract information from the user in a robotic fashion. A typical such system incorporates an NLU component which extracts information from the user query, identifies the type of query and the information required to process that request. Subsequently, a back-end Dialogue Manager (DM) communicates with external servers to execute the user request and decides on the response of the dialogue system to the user. Finally, the response returned by the DM is converted into natural text by the NLG component.\footnote{In this paper the Natural Language Generation refers to text generation.}  
\begin{table}[t!]
\small
\textbf{Dialogue 1:} Task-oriented dialogue via turn-level goals \\
\begin{tabularx}{0.48\textwidth}{|lX|}
\hline
\textbf{User} & Hi Jarvice. I want to buy a movie ticket. \\
\textcolor{gray}{\textit{Turn Goal}} & \textcolor{gray}{\textit{"action", "comedy"}} \\
\textbf{System} & Do you want to see action, or comedy, or some animation? \\
\textbf{User} & I'd like to watch some animation. \\
\hline
\end{tabularx}

\textbf{Dialogue 2:} Transition between chitchat and task-oriented dialogue via turn-level goals\\
\begin{tabularx}{0.48\textwidth}{|lX|}
\hline
\textcolor{gray}{\textit{Turn Goal}} &  \textcolor{gray}{\textit{-}} \\
\textbf{System} & I heard the weather is going to be good this weekend. \\
\textbf{User} & Never mind... I want to buy a movie ticket. \\
\textcolor{gray}{\textit{Turn Goal}} &  \textcolor{gray}{\textit{I'd like that Thai restaurant.}} \\
\textbf{System} & The traffic may be bad out there to go to any movie theater. Can I interest you with a few nearby restaurants, like King of Burgers or Best of Thai, instead? \\
\textbf{User} & OK, I'd like that Thai restaurant. \\
\hline
\end{tabularx}
\caption{Dialogues guided by turn-level goals combining chit-chat and task-oriented dialogue parts. In dialogue 1, when a turn-level goal is provided, the dialogue agent guides the dialogue in a direction talking about the content of the goal while the overall dialogue context is considered. In dialogue 2, when no singe-turn goal is provided, the dialogue agent generates a not goal-oriented system response fitting to the dialogue context by performing social chit-chat.}
\label{tab:example_dialogues}
\end{table}

\noindent The NLG component is one of the most difficult aspects of a task-oriented dialogue system to develop and requires substantial research efforts. This has led most practical implementations of such dialogue systems to implement painstakingly developed pre-determined text responses instead. In addition, these systems can not mould a conversation with a user based on an instantaneous turn-level goal provided by an external source, for example an application to book a train ticket. Instead, most practical dialogue systems require an accurate definition of a dialogue goal before deployment of the dialogue system, enabling no dynamic change of the dialogue goal. 

As shown in this paper, the concept of turn-level  goals enable five dialogue modes: (i) pure chit-chat, (ii) single-domain task-oriented dialogues, (iii) multi-domain task-oriented dialogues with dialogue goal transitions, (iv) transition from chit-chat to task-oriented dialogues as well as (v) transition from task-oriented dialogues to chit-chat. Based on the resulting \testtt{system-user} conversation, one or many oracle components can extract relevant information in order to fulfill tasks. An example provides dialogue 1 in Table \ref{tab:example_dialogues} using the turn-level goal \textit{"action", "comedy"}. This goal guides the conversation about movie tickets seamlessly into the topics \textit{"action"} and \textit{"comedy"}, while the dialogue context is still considered. Using the same concept, chit-chat can be initiated by not providing a turn-level goal as shown in the first turn of dialogue 2 in Table  \ref{tab:example_dialogues}. In contrast, for the second turn a turn-level goal is presented enabling to guide the conversation from chit-chat to a task-oriented dialogue about restaurant booking. 

\noindent Hence, there is a need for \textit{ductile dialogue agents} that can direct conversations based on turn-level goals provided by an external source, and do so in the least robotic sounding way possible. In the following, we present our approach for this.  



\noindent To summarize, our main contribution is an NLG component called \textit{DuctileGPT} that:
\begin{itemize}
\item can be guided by turn-level goals,
\item enables the functionality of seamless transitions between chit-chat, task-oriented dialogues and domains.
\end{itemize}
In contrast to prior work, DuctileGPT is an NLG component that does not rely on a dialogue act represented in a semantic form. Instead, DuctileGPT generates responses by taking into account the dialogue history and the turn-level goal.

\noindent We derive DuctileGPT by reviewing the related work in the field of dialogue systems with a focus on natural language generation for task-oriented and chit-chat conversation. Then, we explain our DuctileGPT model in detail and finally, we evaluate our approach using both automatic and human-based evaluation.

\fi

%% file: paper/related_work.tex
\section{Related Work} \label{sec_related}

\paragraph{Dialogue system types and pipelines.} To best suit the use cases, dialogue systems have developed into distinct types; task-oriented, chit-chat, and question-and-answer are the major ones \cite{survey2020, gao2018neural}. Deployed systems can employ fundamentally different pipelines from one type to another. While mature systems are generally built from a pipeline of modules, end-to-end dialogue systems proposed have been proposed lately \cite{peng2020soloist,hosseini2020simple}, for which there exist no clear module boundaries. Nevertheless, NLG is always at the bottom of the pipeline. \\

\noindent Reinforcement learning (RL) can be used to develop a dialogue policy \cite{li2016deep}. In this framework, both the system response-generating NLG and the simulated user's utterance-generating NLG can be considered a part of the environment for RL. \\

\paragraph{NLG in dialogue systems.} NLG as a standalone unit has a long history of development, starting from early work like Eliza \cite{weizenbaum1966eliza} and PARRY \cite{colby1972turing}. In recent years, NLG learned from  statistical methods have been in focus. A learning-based NLG takes either a retrieval-based approach, selecting an appropriate response from a candidate corpus \cite{xiaoice,henderson2019,Shalyminov20}, or a generation-based one, generating a response using a trained model \cite{serban2016building,serban2017hierarchical,xiaoice,conversational_gpt_huggingface, dinan2018wizard}. Sequence-to-sequence and later transformer-based models have been used   \cite{so2019evolved,budzianowski19,brown2020language} for response retrieval and generation. \\

\paragraph{Large dialog datasets and large dialog generation models.} Large datasets enable large models to produce ever-increasing performance. DialoGPT \cite{dialogpt} adapts GPT-2 \cite{gpt2} to the text-dialogue domain, trained on 147M multi-turn dialogues from Reddit discussion thread. Model sizes range from 117M (small) to 762M (large). Meena is a 2.6B parameter model trained end-to-end on 40B words mined and filtered from public domain social media conversations. We note that both datasets  are for uncontrolled dialogues; large datasets for conversations conditioned on the desired future are not as easily obtained. \\

\paragraph{Text infilling.} Text infilling recovers a missing piece within a long-form text, or more ambitiously predicts some texts that can smoothly blend into and fit the context syntactically and semantically \cite{zhu2019text, inset}. A particularly attractive aspect of the text infilling problem formulation is that a massive dataset can be obtained with little to no annotation effort.  
Prior text-infilling works cover different scenarios, from those in which the missing piece contains only a single token \cite{fedus2018maskgan, zweig2011microsoft}, to those in which  an arbitrary number of tokens and sentences are missing \cite{zhu2019text,inset,liu2019tigs,shih2019xl}.

\paragraph{Automatic NLG quality metric.} While many consider it far from perfect, BLEU is a metric long used for judging the fluency of a generated response \cite{papineni-etal-2002-bleu}. Perplexity was shown to correlate well with a human judgment of sensibleness and specificity \cite{adiwardana2020towards}. Sensibleness tries to cover aspects such as common sense and logical coherence, while specificity covers how relevant sentences are within a dialogue context. 

\paragraph{Recent advances in novel dialogue behaviour.}  
Novel use cases and behaviours beyond the present mature dialogue systems have been proposed. Of note, one is to meaningfully support multi-domain dialogues rather than just a simple divide-and-conquer ensemble \cite{dinan2020second}. To guide a conversation to a different goal mid-flight during a conversation is another new behaviour \cite{tang-etal-2019-target,wu-etal-2019-proactive,liu-etal-2020-towards-conversational,DBLP:conf/aaai/XuWNWC20,zhou2020towards}. In some cases, modules learn such behaviours from carefully constructed and annotated datasets such as DuConv \cite{wu-etal-2019-proactive}.

\if false
\section{Related Work} \label{sec_related}

\noindent Research in automated dialogue systems is often demarcated between task-oriented dialogues and chit-chat. Task-oriented dialogue systems tend to involve heavy engineering of the dialogue policy, `intents and slots' based approach for Natural Language Understanding (NLU), and template based approaches for Natural Language Generation (NLG) \cite{gao2018neural}. In this approach, the dialogue system first tries to identify the task requested by the user and then tries to establish pieces of key information necessary to accomplish that task (i.e. filling in the `slots'), either by analyzing the initial prompt by the user or by asking further questions. As an example, a task-oriented dialogue system may be prompted by the user to order takeaway delivery from a particular restaurant, and as a response, the dialogue system may ask which items from the menu the user would like and where the food should be delivered. 

\noindent Despite their success, most such systems extract information from the user in a robotic fashion. A typical such system incorporates an NLU component which extracts information from the user query, identifies the type of query and the information required to process that request. Subsequently, a back-end Dialogue Manager (DM) communicates with external servers to execute the user request and decides on the response of the dialogue system to the user. Finally, the response returned by the DM is converted into natural text by the NLG component.\footnote{In this paper the Natural Language Generation refers to text generation.}  
\begin{table}[t!]
\small
\textbf{Dialogue 1:} Task-oriented dialogue via turn-level goals \\
\begin{tabularx}{0.48\textwidth}{|lX|}
\hline
\textbf{User} & Hi Jarvice. I want to buy a movie ticket. \\
\textcolor{gray}{\textit{Turn Goal}} & \textcolor{gray}{\textit{"action", "comedy"}} \\
\textbf{System} & Do you want to see action, or comedy, or some animation? \\
\textbf{User} & I'd like to watch some animation. \\
\hline
\end{tabularx}

\textbf{Dialogue 2:} Transition between chitchat and task-oriented dialogue via turn-level goals\\
\begin{tabularx}{0.48\textwidth}{|lX|}
\hline
\textcolor{gray}{\textit{Turn Goal}} &  \textcolor{gray}{\textit{-}} \\
\textbf{System} & I heard the weather is going to be good this weekend. \\
\textbf{User} & Never mind... I want to buy a movie ticket. \\
\textcolor{gray}{\textit{Turn Goal}} &  \textcolor{gray}{\textit{I'd like that Thai restaurant.}} \\
\textbf{System} & The traffic may be bad out there to go to any movie theater. Can I interest you with a few nearby restaurants, like King of Burgers or Best of Thai, instead? \\
\textbf{User} & OK, I'd like that Thai restaurant. \\
\hline
\end{tabularx}
\caption{Dialogues guided by turn-level goals combining chit-chat and task-oriented dialogue parts. In dialogue 1, when a turn-level goal is provided, the dialogue agent guides the dialogue in a direction talking about the content of the goal while the overall dialogue context is considered. In dialogue 2, when no singe-turn goal is provided, the dialogue agent generates a not goal-oriented system response fitting to the dialogue context by performing social chit-chat.}
\label{tab:example_dialogues}
\end{table}

\noindent The NLG component is one of the most difficult aspects of a task-oriented dialogue system to develop and requires substantial research efforts. This has led most practical implementations of such dialogue systems to implement painstakingly developed pre-determined text responses instead. In addition, these systems can not mould a conversation with a user based on an instantaneous turn-level goal provided by an external source, for example an application to book a train ticket. Instead, most practical dialogue systems require an accurate definition of a dialogue goal before deployment of the dialogue system, enabling no dynamic change of the dialogue goal. 


\noindent Many early examples of neural network-based dialogue systems use seq-to-seq models \cite{sutskever2014sequence} based on recurrent neural networks (RNNs). One such notable example is the Hierarchical Encoder-Decoder Model \cite{serban2016building,serban2017hierarchical} which utilizes multiple RNNs with Long Short-Term Memory cells (LSTM) \cite{LSTM}  that embed text into fixed-size feature vectors at various 'levels', such as at word, sentence or paragraph level. 
The application of the attention mechanism to seq-to-seq models \cite{bahdanau2014neural} improved their ability  to capture long-term dependencies. 
The Transformer architecture \cite{vaswani2017attention} dispensed with recurrence, relying on feed-forward layers and attention, builds a computationally efficient encoder-decoder model. Transformer-based models can be trained on huge corpora, such as Wikitext. This has led to the development of pre-trained general purpose language representation models \cite{Bert, gpt2, gpt} that can be adapted to downstream tasks through low-resource fine-tuning. In particular, BERT \cite{Bert} and GPT \cite{gpt}, trained in a self-supervised fashion and therefore not requiring labelled data, have shown great promise. 

\noindent Since dialogue datasets are typically small, the availability of pre-trained multi-purpose models is particularly relevant for dialogue systems. DialoGPT \cite{dialogpt} adapts GPT-2 \cite{gpt2} to response generation, using a corpus of 147M Reddit comment chains. This extensive pre-training on a dialogue-like data set allowed DialoGPT to achieve state-of-the-art in automatic and human evaluation results on smaller dialogue datasets after fine-tuning. Recently, GPT-3 \cite{brown2020language} was released, with improved benchmark results over GPT-2. However, GPT-3 is licensed exclusively to Microsoft and access is only possible via a paid cloud API. 
\noindent Most such models require fine-tuning on training data from the target task. Therefore, there was a need to construct labelled training data for research on dialogue systems. Consequently, various challenges such as DSTC-8 \cite{dstc8} and conferences \cite{dinan2020second} were launched with sole focus on dialogue systems, which led to various labelled public datasets \cite{dstc8, budzianowski2018multiwoz, byrne2019taskmaster}. 

\noindent Most dialogue systems can be broadly categorized into three types: 'question answering', 'open domain chit-chat' and 'task oriented' dialogue systems. Going forward, we will give a brief overview of literature with relevance to our work in chit-chat and task-oriented dialogue systems, while for question answering we refer to the literature \cite{tay2018hyperbolic,zhong2019coarse,bordes2015large, yang2019end,joshi2019spanbert, yang2019xlnet, le2020self}.


\noindent
\paragraph{Open-domain chit-chat: } The RNN based XiaoIce \cite{xiaoice} chatbot and its derivatives rose to prominence rapidly, being deployed extensively on popular platforms including Twitter and WeChat, serving millions of users. Transformer-based models (such as GPT-2) have been used to both generate general-purpose text and/or conversations \cite{conversational_gpt_huggingface, dinan2018wizard} as well as in more specialized applications including serving as an RPG (role-playing game) game master \cite{aidungeon}, generating speeches in the styles of UK political figures \cite{gpt2parliamentaryspeeches} or even writing poetry \cite{gpt2poetry}. 

\noindent
\paragraph{Task-oriented dialogue systems:} 
Conventional task-oriented systems are multi-modular and the application of pre-trained models to this field is still in its inception. BERT-DST \cite{chao2019bert} applies BERT to the dialogues state tracker (DST), the module that keeps track of the intention of the users, to make the dialogue dynamic by inferring the slots to be filled from the context. BERT has also been applied to pre-trained response selection for a system based on information-retrieval \cite{henderson2019}. GPT-2 has shown potential in language generation for task-oriented applications, through fine-tuning only \cite{budzianowski19} as well as with an additional component performing information retrieval from a support dataset \cite{Shalyminov20}. In addition, transformer based architectures have also been applied to guide dialogue policy \cite{vlasov2019dialogue}. In the most recent works, end-to-end task-oriented dialogue systems evolved based on transformer architectures \cite{peng2020soloist, hosseini2020simple}. SOLOIST is an auto-regressive model predicting the dialogue state and system response based on the dialogue history, the previous dialogue state and the database state \cite{peng2020soloist}. SimpleTOD is a similar approach generating additional dialogue actions -- like inform or request -- leading to improved results \cite{hosseini2020simple}.
However, the present work requires the dialogue state representation to be carefully designed and all slot keys of the dialogue state being available at the beginning of a conversation. 

\noindent To summarise the related work, models are either trained purely on chit-chat, making it impossible to control the dialogue direction, or they are trained to fulfil a task-oriented dialogue but cannot react on instantaneous turn-level goals or are widely limited in their chit-chat capabilities. In the following sections, we explain how our proposed novel model, DuctileGPT,  overcomes this gap.

\fi

%% file: paper/ductiletransformer.tex
\section{Methods}
In a dialogue system, the NLG is at the bottom of the stack. The components above the NLG can be very complicated. As these components are outside of the scope of this paper, we will just refer to them by the ``top of the stack". Loosely speaking, we treat the top of the stack as a function that projects a future to an FBNLG to bridge to. Additionally noted, in the rest of the paper we discuss how FBNLG is used to generate a system response; when used as a dialogue simulator, it might be necessary to swap the notation for user and system utterances.\\ 

\noindent We note that an FBNLG learned in the fashion given below is only distributionally correct. In other words, an FBNLG tries only to maintain that the generated system response fits well with the distribution of the dataset it is trained on. To be useful in a real-world scenario, a generated dialogue has to be grounded in facts. Nevertheless, we think that factual grounding techniques can be added to an FBNLG in a way that is orthogonal to the main point of this paper. We leave factual grounding outside of our paper.

\subsection{Future projection and future bridging}

\begin{figure}[tb]
\includegraphics[width=0.48\textwidth]{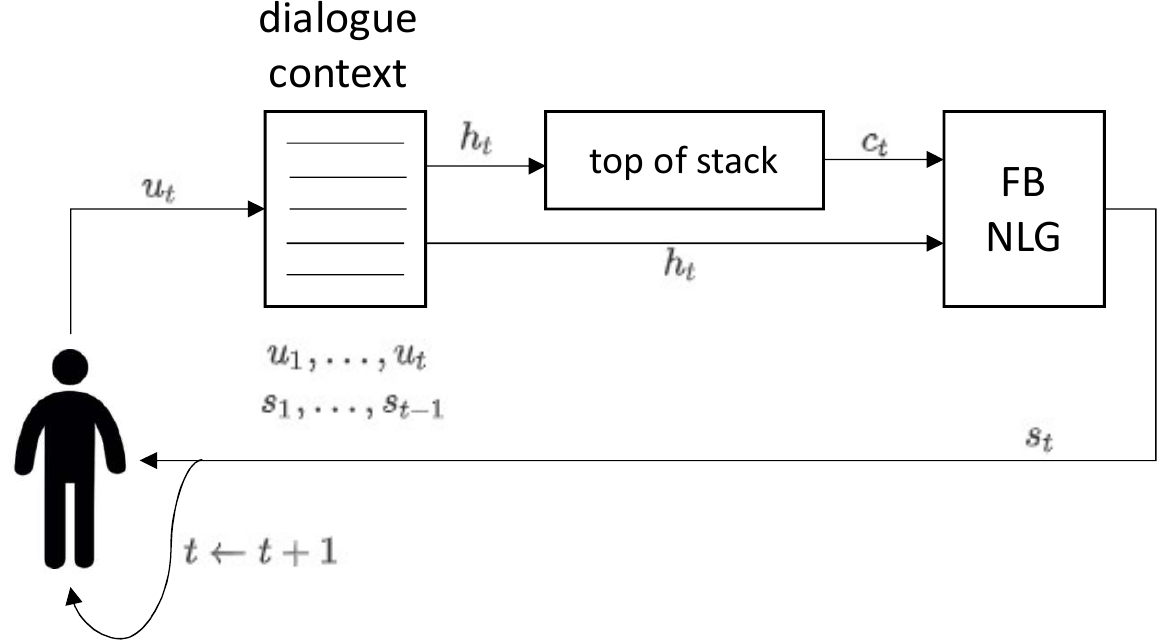}
\caption{Illustration of future bridging and future projection. $u_t$ and $s_t$ are the user's and system's utterance at turn $t$, respectively. $c_t$ is the desired future.} 
\label{fig:DuctileGPT}
\end{figure}

We define the act of \textit{future bridging} and the associated act of \textit{future projection} as follows. At turn $t$, the dialogue context $h_t$ consists of user utterances $u_1$, $...$, $u_t$ and system responses $s_1$, $...$, $s_{t-1}$. By future projection, the top of stack proposes a desirable future (user or system) utterance at $\delta$ turns away, $c_{t}$ = $u_{t+\delta}$ or $s_{t+\delta}$, to the NLG. The NLG attempts to bridge from the context to $c_{t}$, thus producing the immediate system utterance $s_t$ as result. This NLG behaviour is called future bridging. After the user receives $s_t$ and returns with $u_{t+1}$, the cycle repeats. This is shown in Figure \ref{fig:DuctileGPT}. \\

\noindent We can view future bridging as a text infilling problem.  Specifically, dialogue future bridging is to learn the probability of the immediate system response 
$s_t$ which, together with a trailing trajectory $\tau_{s_t}$ = $u_{t+1}$, $s_{t+1}$, $...$,  fills between the left-sided context $h_t$ and the right-sided context $c_t$: 
\begin{multline}
P(s_t|u_1,s_1, ..., u_t, c_{t}) \\
= \sum\limits_{\tau_{s_t}} P(s_t, \tau_{s_t}|u_1,s_1, ..., u_t, c_{t}).  
\label{eq:joint_prob_duct}
\end{multline}

\subsection{Preparation of dataset}

\noindent Here we explain how future bridging allows an FBNLG to be trained in a self supervised manner decoupled from the rest of the dialogue system. \\

\noindent Start with a large collection of dialogues, $\mathcal{D}$ = $ \left\{ d^1, ..., d^N \right\} $. Each dialogue $d^i$ consists of alternating user turns $u^i_t$ and system turns $s^i_t$. Assuming that the projected future is to be a user utterance $\delta$ turns away. We can construct a sample $x^j$ for the self-supervised training set $\mathcal{X}$ from $\mathcal{D}$ by sampling a dialogue $d^i$, taking the context $h^i_t$ up to time step $t$, the future user utterance at the target time step $u^i_{t+\delta}$, the immediate system response $s_t^i$ to bridge to that future, and optionally some negative examples $s^{\overline{i}}_{t}$ which are systems' responses randomly sampled from other dialogues of $\mathcal{D}$. Specifically, $\mathcal{X}$ is made up of 
\begin{equation}
x^j = (h^i_t, c^i_t; s^i_t; s^{\overline{i}}_{t}).   
\end{equation}

\noindent For the purpose of this paper, we only consider the top of stack to be a mapping $f_p$ from a dialogue context $h_t$ and a situational context $b_t$ to a desired future user utterance $u_{t+\delta}$; $f_p: \mathcal{H} \times \mathcal{B} \rightarrow \mathcal{U}$. A situational context contains real-time application-specific incentives such as inventory level. This mapping is learned from a dataset $\mathcal{Z}$ that is typically collected over the particular use case of the application. A small subset of $\mathcal{Z}$ can be optionally withheld to produce a future bridging dataset $\mathcal{X}'$ in the aforementioned manner. This $\mathcal{X}'$  can be used to adapt an FBNLG to the top of stack in a few-shot manner. \\

\noindent The self-supervised pre-training and task-specific zero- or few-shot adaptation aspects of the proposed FBNLG follows exactly that of the transformer-based language model paradigm. We expect that FBNLG shall exhibit similarly higher performance and superior downstream versatility due to the massive data advantage. 

\subsection{Training}


\noindent During training, samples from the training set $\mathcal{X}$ are used to train an FBNLG. An important special case is that, in some situations, the top of stack may signal that it is content with letting the dialogue proceed spontaneously. To model these situations, one can randomly take a subset of training samples and replace the target futures of these samples by NULL. The NULL substitution can be done statically at dataset construction time or dynamically at training time. \\ 

\noindent The goal of FBNLG is to learn the conditional utterance probability in Equation (\ref{eq:joint_prob_duct}). An FBNLG can be either generative or retrieval-based. For an autoregressive generator, the goal can be further expressed as, 
\begin{equation}
p(s_t|h_t, c_t) = \prod_{k=0} p(s_{t,k}|h_t, c_t, s_{t,1:k-1})\text{,}
\label{eq:token_prob_duct}
\end{equation}
with $k$ denoting an index over the tokens of $s_t$ in causal order, and $s_{t,1:k-1}$ denotes the generated tokens 1 through $k-1$.\\

\noindent Following originally presented in \cite{bengio2003neural}, we propose to use an adapted version of the language modelling objective. We optimize for a joint objective of \textit{response generation} (RG) and \textit{response selection}:
\begin{equation}
    \mathcal{L}(\mathcal{X}) = \sum_j^{|X|} \alpha_{\text{RG}} \mathcal{L}_\text{RG}(x^j) + \alpha_{\text{RS}} \mathcal{L}_\text{RS}(x^j)
\end{equation}

\noindent The response generation objective, $\mathcal{L}_{\text{RG}}$, measures the log probability of positive samples. For an autoregressive NLG, the response generation objective for sample $x^j$ is computed as follows: 
\begin{equation}
\begin{aligned}
\mathcal{L}_{\text{RG}}(x^j) = \sum_{k=1} \log p(s^j_{t,k}| h_{t}^j, c_{t}^j, s_{t,1:k-1}^j). \\
\end{aligned}
\end{equation}

\noindent The response selection objective $\mathcal{L}_{\text{RS}}$ measures the rejection of negative samples. The use of the response selection objective improves the overall language model performance as shown in prior work \cite{peng2020soloist}. Following \cite{peng2020soloist}, from a training sample $x^j$, we randomly present either $(h_t, c_t, s_t)$ or $(h_t, c_t, \overline{s}_t)$ to the NLG, asking it to classify whether the correct response ($y$=1) or a distracting response ($y$=0) is present. The objective is computed via binary cross-entropy loss as follows:
\begin{equation}
\begin{aligned}
\mathcal{L}_{\text{RS}}(x^j) = &y \log p(h_t^j; c_t^j; s_t^j) \\ &+ (1-y) \log (1-p(h_t^j; c_t^j ; \overline{s}_t^j)).
\end{aligned}
\end{equation}

%% file: paper/evaluation.tex
\section{Evaluations}


\noindent We built a prototype FBNLG and evaluated the quality of generated dialogues via automatic and human-based evaluations. We furthermore experimented goal transition within a task-oriented conversation, and dynamic dialogue type switching from one to the other.
\footnote{A demonstration of this FBNLG prototype can be found under \url{http://www.github.com/ANONYMOUS}.}

\subsection{Model Architecture, Initialization, and Training}

Ideally, one shall pre-train a text infiller from scratch for an FBNLG. However, this pre-training requires a very significant computational facility. In our evaluation, we instead built a prototype from an openly available pre-trained language model, DialoGPT. We selected the small version of DialoGPT (117M parameters) as the starting point. Later, we fine-tuned the starting point on the modest datasets in Table \ref{tab:datasets} for two effects. One, to convert the auto-regressive DialoGPT model to a text infilling model; and two, to tilt the data distribution from uncontrolled dialogues to conversations conditioned on desired futures.  We note that \textbf{MultiWOZ} \cite{budzianowski2018multiwoz}, \textbf{Schema-Guided Dialogue} \cite{rastogi2019towards}, and \textbf{Taskmaster-1} \cite{byrne2019taskmaster} are datasets for task-oriented type, and \textbf{DailyDialog} \cite{li2017dailydialog} for chit-chat type. By training with  both task-oriented and chit-chat datasets, an FBNLG learns to fulfil a purpose when given a meaningful future, or continue the ongoing dialogue with small talks when given a NULL future. \\

\begin{table*}
\centering
\small
\begin{tabular}{lllll}

\hline \textbf{Stats}                   &  \textbf{MultiWOZ} &  \textbf{SGD}     &  \textbf{Taskmaster} &  \textbf{DailyDialog} \\ \hline
\# domains             & 7        & 16      & 6          & n/a         \\ 
\# dialogues           & 8,438    & 16,142  & 13,215     & 13,118      \\ 
Total no. of turns      & 113,556  & 329,964 & 303,066    & 103632      \\ 
Avg. turns per dialogue & 13.46    & 20.44   & 22.9       & 7.9 \\ \hline
\end{tabular}
\caption{Statistics for the fine-tuning dataset}
\label{tab:datasets}
\end{table*}

\noindent We prepared the self-supervised datasets according to the procedure given in the previous Section. From a chit-chat source dataset, the projected future $c_t$ is set to NULL. On the other hand, from a task-oriented source dataset, we use the next user utterance as the future. In other words, $c_t \leftarrow u_{t+\delta}$, with $\delta = 1$.  

\subsection{Experimental Setup}
\begin{table*}[t]
\centering
\small
\begin{tabular}{l|rrrrrrrrrr}
\hline
\multicolumn{1}{c|}{\multirow{2}{*}{Model}} & \multicolumn{2}{c}{SGD}      & \multicolumn{2}{c}{SGD single} & \multicolumn{2}{c}{SGD multi} & \multicolumn{2}{c}{MultiWOZ} & \multicolumn{2}{c}{DailyDialog} \\
\cline{2-11}
\multicolumn{1}{c|}{}                                 & BLEU↑          & PPL↓        & BLEU↑             & PPL↓            & BLEU↑             & PPL↓           & BLEU↑          & PPL↓        & BLEU↑           & PPL↓          \\
\hline
Baseline                                             & 14.5          & 56          & 13.8             & 61              & 14.7            & 57             & 20            & 34          & 1.4           & 385           \\
Model D                 & 19           & \textbf{23} & 18             & \textbf{26}     & 19.2             & \textbf{22}    & 21.3          & \textbf{23}          & 1.8           & \textbf{316} \\
Main prototype                  & \textbf{23.1} & 24          & \textbf{22}     & 28     & \textbf{23.3}    & 23             & \textbf{24.4} & 32 & \textbf{2}   & 595           \\
\hline
\end{tabular}
\caption{Automatic Evaluation Results. \textit{Baseline} is the baseline model that has not learned future bridging. \textit{Model D} is a FBNLG model suffering from some distribution mismatch. }
\label{tab:auto_eval}
\end{table*}

The main prototype is built in a two-step process. From a DialoGPT model, we first additionally pre-train it on the relatively larger SGD and Taskmaster to learn future bridging. Subsequently, we adapt it to the target domains by fine-tuning it with DailyDialog, SGD, and MultiWOZ. We compare the main prototype against two benchmark models for ablation studies. The baseline model is a DialoGPT model directly fine-tuned on DailyDialog, SGD and MultiWOZ using the all-NULL future bridging objective. It generates system response only from the context. Obviously, this NLG is of little use; it serves to establish the value of future bridging. The other model (model D) is a DialoGPT model directly fine-tuned on DailyDialog, SGD and MultiWOZ using proper future bridging objective. This model suffers from distribution mismatch between the data it is pre-trained on and that of the domain of interest. It allows us to estimate the value of a large scale, unlabelled oral dialogue dataset in future bridging. \\

\begin{table*}[t]
\centering
\small
\begin{tabular}{l|l|l}

\hline 
 \textbf{Dialogue type transition} & \textbf{Specificity} &  \textbf{Sensibleness}  \\ \hline
 chit-chat to task-oriented & 4.09	& 5.33 \\
 task-oriented to chit-chat & 4.98	& 5.46 \\
\hline 
\end{tabular}
\caption{Human-based evaluation for dialogue-type transitions.}
\label{tab:human_eval}
\end{table*}

\begin{table*}[t]
\centering
\small
\begin{tabular}{l|l|l}

\hline 
 \textbf{Dialogue type transition} & \textbf{Specificity} &  \textbf{Sensibleness}  \\ \hline
 chit-chat to task-oriented & 4.16	& 5.03 \\
 task-oriented to chit-chat & 4.98	& 5.46 \\
\hline 
\end{tabular}
\caption{Human-based evaluation for future-bridging performance}
\label{tab:human_eval_guidance}
\end{table*}

\if false
\begin{table*}[t]
\centering
\small
\begin{tabularx}{0.99\textwidth}{l|llXX|cc}

\hline \textbf{ID} & \textbf{Dialogue type transition}                   & \textbf{Model} & \textbf{Future-projection from DM} & \textbf{Future-projection shown to turk}  & \textbf{Specificity} &  \textbf{Sensibleness}  \\ \hline
1 & Chit. to Task. & Main prototype & X & X & 4.16	& 5.03 \\
2 & Chit. to Task. & Main prototype & X & - & 4.09	& 5.33 \\
3 & Chit. to Task. & Baseline & - & - & 4.40	& 5.20 \\ \hline
4 & Task. to Chit. & Main prototype &-& - & 4.98	& 5.46 \\
5 & Task. to Chit. & Baseline & - & - & 4.98	& 5.31 \\ \hline

\end{tabularx}
\caption{Human-based evaluation. Apart from the baselines, all models have been trained including future projection. In 1, the turks had knowledge about the projected future; in 2, the projected future is hidden to the turks and 3 is the baseline. In 4 and 5 both models have been applied without future-projection during inference.}
\label{tab:human_eval}
\end{table*}
\fi
\noindent We performed both automatic and human evaluations. We compute the BLEU score and perplexity on held-out test datasets from SGD, MultiWOZ and DailyDialog \cite{papineni-etal-2002-bleu}. Furthermore, for the two novel behaviours we selected - dynamic dialogue type switching and dynamic dialogue goal transition - We carried out a human evaluation study using crowd workers. While limited in scale by various practical factors, human evaluation is the true measure on dialogue generation quality. Human-evaluation details are given in the Appendix. 



\subsection{Dialogue Quality Results}

\noindent First, we checked whether FBNLG can perform chit-chat type dialogues. We tested the main prototype and the two baseline models against the test dataset of DailyDialog with automatic evaluation. Results are given in Column \textit{DailyDialog} in Table \ref{tab:auto_eval}. Primarily because in this evaluation the top of stack does not give a future to bridge to, these models all score poorly on both scales. Nevertheless, by our own examinations, we think that one shall not jump to the conclusion that the generated dialogues are of low quality simply from the low automatic scores. As is shown in the human evaluation results below, for generated dialogues that contain chit-chat portions, human evaluation actually rated the sensibleness and specificity with a satisfactory rating on the Likert scale. \\

\noindent Next, we check whether FBNLG can be used in a single-domain task-oriented dialogue. We tested the three models 
against the single-domain dialogues from the SGD test dataset. The results can be found in Table \ref{tab:auto_eval} under the column "SGD single". Comparing Model D over the Baseline, the use of future bridging dramatically improves the BLEU score by around 4 as well as cuts the perplexity by a factor of nearly 3. Comparing the main prototype over Model D, we can see that pre-training against a modest-sized oral dialogue dataset can furthermore improve the fluency by 4 BLEU points. \\

\noindent We then tested the three models against the multi-domain dialogues from the SGD test dataset assuming an oracle top of stack. Such a dialogue involves challenging domain transition within a single conversation. While the single-domain automatic evaluation results are good, the multi-domain results are no less encouraging. Automatic evaluation results for multi-domain task-oriented dialogues can be found in Table \ref{tab:auto_eval} in column "SGD multi". Here again we observed that the use of future bridging improves the BLEU by 4 and reduces the perplexity by a factor of nearly 3. Similarly, once the text-oriented dialogue distribution bias of DialoGPT is partially compensated, the BLEU score of the main prototype rises by 4. \\

\noindent We proceeded to test the viability of FBNLG over a chosen novel dialogue system behaviour, dialogue type transition. Because we could not find any existing dataset for this purpose, we created a small test-only dataset. In this dataset, we have both chit-chat-to-task-oriented, and task-oriented-to-chit-chat dialogues. We carefully maintained the nature of the chit-chat dialogues similar to DailyDialog, and the task-oriented dialogues similar to the aforementioned task-oriented datasets. Example dialogues are given in Table \ref{tab:example_ducttransf}. \\

\noindent Dialogue 1 in Table \ref{tab:example_ducttransf} is an example of a dialogue starting from chit-chat about the city of New York and then switch to a task-oriented real estate promotion. Throughout the chit-chat portion, the top of stack keeps giving NULL as the future. Then, it gives a purposeful future for the model to bridge to. The quality of the generated system response is measured. Dialogue 2 is an example in which the conversation switches from task-oriented to chit-chat. Purposeful futures are given during the task-oriented portion of the dialogue. Toward the end of the dialogue, the top of stack stops gives a NULL future. The FBNLG is expected to produce an appropriate chit-chat response. The quality of this response is measured. \\ 

\noindent We asked crowd workers to rate the specificity and sensibleness of the system response generated by the main prototype corresponding to the dialogue type-switching turn on the Licker scale. Note that our main prototype has never been trained on a dialogue-type transition dataset. From this perspective, what was evaluated was not only whether FBNLG can learn to support never-planned behaviour. It is, indeed, the \textbf{zero-shot} capability of FBNLG to support a new behaviour on the fly. The human evaluation results of dialogue type transitions can be found in Table \ref{tab:human_eval}. Here we observe that both sensibleness of and specificity are satisfactory to the crowd workers. In summary, the experiments of dialogue type transition points to the potential of a pre-trained FBNLG being universal, able to support present and never-planned behaviours.   

\begin{table*}[t!]
\small
\centering
\textbf{Dialogue 1:}  Chit-chat type to task-oriented type \\
\begin{tabularx}{0.98\textwidth}{|p{2.5cm}X|}
\hline
\textbf{User} &  Isn't it wonderful walking here? \\
\textbf{System} & What do you mean? \\
\textbf{User} &  I mean look at all these magnificent buildings around us. \\
\textbf{System} & Yes, look over there. That's the Empire State Building. My book says it's 102 stories tall. \\
\textbf{User} &  It's quite famous but don't you think it looks a bit old-fashioned? \\
\textbf{System} & You're right, but when it was built in 1930 it was a marvel of technology and engineering. \\
\textbf{User} &  What other important buildings are we going to see on Fifth Avenue?\\
\hline
\textcolor{gray}{\textit{Projected future}} & \textcolor{gray}{\textit{I'm looking for a house to stay in.}}\\
\textbf{FBNLG response} & Well, there's a really nice place to live in. It's called the Palace Hotel, and it has a great view of the city. I think you'll love it!\\
\hline
\textbf{Baseline response}  & There's a museum called the Great Wall of China. It's very famous. You should go see it!\\
\hline
\end{tabularx}

\vspace{0.2em}

\begin{tabularx}{0.98\textwidth}{|p{2.5cm}X|}
\multicolumn{2}{c}{\textbf{Dialogue 2:} Task-oriented type to chit-chat type} \\
\hline
\textbf{User} & I am looking for some child friendly attractions in Nairobi. \\
\textbf{System} & How about a musical called Lions Kings?\\
\textbf{User} & That sounds great.\\
\hline
\textcolor{gray}{\textit{Projected future}} & \textcolor{gray}{\textit{NULL}} \\
\textbf{FBNLG response} & Is there anything else I can help you with?\\
\hline
\end{tabularx}

\caption{Example of dialogue type transition within a single conversation. Dialogue 1: chit-chat to task-oriented. Dialogue 2: task-oriented to chit-chat.}
\label{tab:example_ducttransf}
\end{table*}

\subsection{Future Bridging Results}

The results above indicated that FBNLG can generate system responses that are reasonably fluent, specific, sensible, even for never-planned dialogue behaviours without any learning when viewed as a continuation of the dialogue context. But does an FBNLG actually produce a result that fulfils its future bridging obligation? Intuitively, one could evaluate the future bridging quality by evaluating the quality of the overall dialogue, spanning from the left-side context $u_1$, $s_1$, ..., $u_t$, through the generated response $s_t$, to the right-side context $c_t$ = $u_{t+1}$. A higher dialogue quality should be positively correlated with the future bridging performance. \\ 

\noindent We note that the subtle difference from the evaluations in Section 4.3 is that, while previously tests are conducted to  evaluate the quality of $s_t$ given only the left-side context, here one makes evaluations taking into account the right-side context $c_t$ = $u_{t+1}$ as well. \\

\noindent Using human-based evaluations, we asked the crowd workers to evaluate the specificity and sensibleness of the generated dialogue. The results are shown in Table \ref{tab:human_eval_guidance}. We see that both the sensibleness and the specificity are acceptable, indicating that the FBNLG indeed largely fulfils the future bridging objective. \\

\noindent By comparing Table \ref{tab:human_eval} and \ref{tab:human_eval_guidance}, one might draw the conclusion that the causal quality evaluation in Table \ref{tab:human_eval} can be a good proxy score for future bridging performance. However, in our opinion, whether this is incidental requires some further study. \\